\pdfoutput=1

\documentclass[11pt]{article}
\usepackage[table]{xcolor}
\usepackage[]{acl}

\usepackage{times}
\usepackage{latexsym}

\usepackage{amssymb}
\usepackage{amsmath} 
\usepackage{booktabs} 
\usepackage{array}
\usepackage{graphicx} 
\usepackage{multirow}
\usepackage{ulem}
\usepackage{makecell} 
\usepackage{float}

\usepackage{colortbl}

\usepackage[T1]{fontenc}

\usepackage[utf8]{inputenc}

\usepackage{microtype}

\usepackage{inconsolata}

\title{Rational Sensibility: LLM Enhanced Empathetic Response Generation Guided by Self-presentation Theory}

\author{Linzhuang Sun$^{1,3}$, \ Yao Dong$^{2}$, \ Nan Xu$^{2}$\thanks{\quad Corresponding author, nan.xu@wenge.com} , \ Jingxuan Wei$^{1,3}$, \ Bihui Yu$^{1,3}$, \ Yin Luo$^{2}$   \\
  $^{1}$Shenyang Institute of Computing Technology, Chinese Academy of Sciences\\
  $^{2}$Beijing Wenge Technology Co., Ltd 
  $^{3}$University of Chinese Academy of Sciences \\
  \texttt{sunlinzhuang21@mails.ucas.ac.cn} \\
  }

\begin{document}

\maketitle
\begin{abstract}
The development of Large Language Models (LLMs) provides human-centered Artificial General Intelligence (AGI) with a glimmer of hope. Empathy serves as a key emotional attribute of humanity, playing an irreplaceable role in human-centered AGI. Despite numerous researches aim to improve the cognitive empathy of models by incorporating external knowledge, there has been limited attention on the sensibility and rationality of the conversation itself, which are vital components of the empathy. However, the rationality information within the conversation is restricted, and previous methods of extending knowledge are subject to semantic conflict and single-role view. In this paper, we design an innovative encoder module inspired by self-presentation theory in sociology, which specifically processes sensibility and rationality sentences in dialogues. And we employ a LLM as a rational brain to decipher profound logical information preserved within the conversation, which assists our model in assessing the balance between sensibility and rationality to produce high-quality empathetic response. Experimental results demonstrate that our model outperforms other methods in both automatic and human evaluations.

\end{abstract}

\section{Introduction}

Empathetic response generation, the capacity to perceive emotion of individuals and response accordingly, is integral in the pursuit of intelligent agents\cite{hofmann2010empirical}. 
In psychology, sensibility and rationality are crucial components of an individual's empathy.
Lack of sensibility makes it ineffectual to emotionally relate to users. And conversely, the absence of rationality may result in emotional empathy and the symptom of unmitigated communion\cite{fritz1998distinctions}. However, rational sensibility, known as cognitive empathy, allows for a better comprehension of users while decreasing one's own negative emotional experience\cite{smith2006cognitive}. 

Although many researchers endeavor to enhance the cognitive ability of models through external knowledge\cite{ghosal2020cosmic, zhou2021probing, sabour2022cem}, limited attention has been directed towards the inherent expression within the conversation, such as sensibility and rationality. 
Besides, the rational thinking of the dialog itself provides limited information and cannot furnish in-depth perception, such as intention and purpose.
\begin{figure}
\centering 
\includegraphics[width=0.5\textwidth]{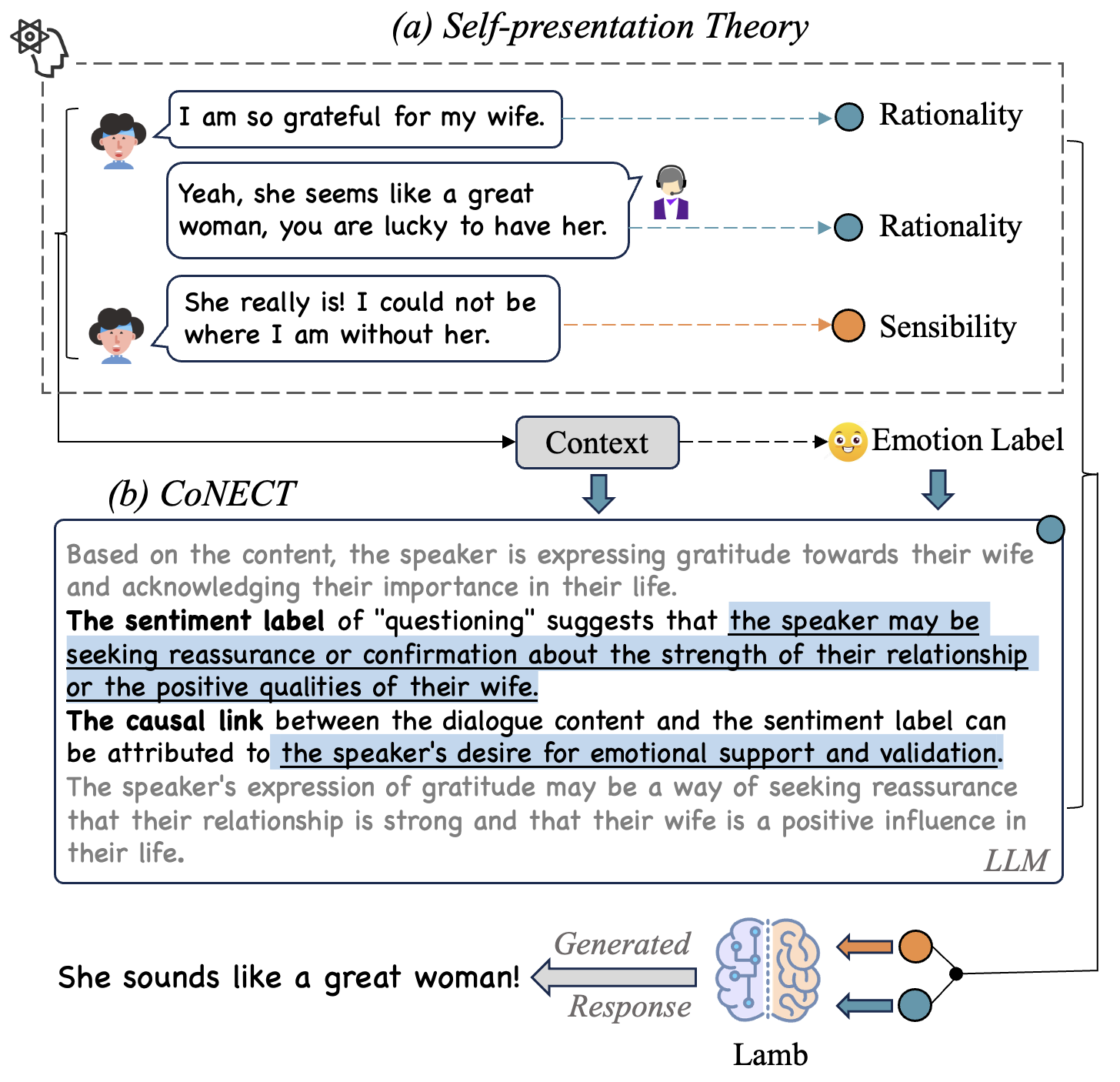} 
\caption{An example of empathetic response from EMPATHETICDIALOGUES dataset. (a) According to the sociological theory of self-presentation, categorizing conversation utterances into sensibility and rationality groups. (b) the chain of emotion-aware empathetic prompting based on LLM. }
\label{Fig.demo}
\end{figure}
Previous methods usually adopt COMET\cite{bosselut2019comet} to enhance the commonsense reasoning ability of models, which is constructed on two large knowledge graphs, ATOMIC\citep{sap2019atomic} and ConceptNet\citep{speer2017conceptnet}, including a huge amount of social commonsense knowledge. However, this method exists two limitations: semantic conflict and single-role view.
Commonsense knowledge can be divided into five categories: intent, need, effect, want, and react. These categories are relatively independent and difficult to ensure semantic consistency, which is the main reason for semantic conflict\cite{cai2023improving}. Furthermore, previous methods are limited to focusing on a single-turn utterance to expand external knowledge with a single-role view, e.g., speaker or listener. To tackle the above issues, we seek solutions based on the intersection of sociology and the large language model (LLM).

In sociology, self-presentation theory\cite{baumeister1987self, jensen2003we} divides sentences into categories: sensibility and rationality. In sensibility, the emotionally charged utterances are used to accentuate the individual's personality traits, depict personal state and project desired character. In rationality, sentences primarily serve the purpose of transmitting information, expressing opinions, and engaging in logical thought processes in an objective manner. As illustrated in Figure~\ref{Fig.demo}(a), the dialogue is divided into sensibility and rationality sentences with the aim to achieve more accurate context encoding. 

Furthermore, we employ the LLM as a rational brain to provide logical reasoning for empathetic response without limitation of semantic conflict and single-role view. Particularly, we design an innovative module, rational representation enhancement, which are mainly composed by \textbf{C}hain \textbf{o}f Emotio\textbf{n}-awar\textbf{e} Empatheti\textbf{c} promp\textbf{t}ing(CoNECT). As shown in Figure~\ref{Fig.demo}(b), CoNECT employs an affective indicator to facilitate the evaluation and correlation of contextual connections, based on the extensive knowledge within LLM. This approach enables empathetic reasoning and boasts notable advantages: 
1) \textbf{Rational reasoning.} Examine the portrayal of emotional labels within historical context and analyze the intrinsic psychological expression. 2) 
\textbf{Multi-role view.} Offer a comprehensive perspective when perceiving the emotions and intentions from the view of speaker and listener on a global scale.

In this paper, we present \textbf{L}LM enhanced emp\textbf{a}thetic response generation \textbf{m}odel guided \textbf{b}y self-presentation theory, \textbf{Lamb}. The model consists of three modules: 1) Rational-sensible Encoder.
Emotionally charged sentences are identified by the pretrained emotion-cause model. Then, we utilize a designed attention mechanism to combine the sensible and rational sentences. 2) Rational Representation Enhancement.
In the context of a conversation and the corresponding emotion label, we employ the CoNECT and COMET methodologies to generate rational knowledge for Lamb. 3) Rational-sensible Decoder.
We employ a cross-attention mechanism that allows the decoder to perceive context, commonsense knowledge and CoNECT data.

Our main contributions are listed as follows:
\begin{itemize}

\item Guided by self-presentation theory, we focus on the sensible and rational expression of the conversation itself to enhance cognitive empathy.

\item We introduce the chain of emotion-aware empathetic data into the empathetic response generation task, which provides multi-role aware rational reasoning.

\item Our model, Lamb, uses weighted context and external knowledge to understand speaker's emotional state and generate suitable empathetic response.

\item Experiments demonstrate that Lamb generates more empathetic response compared with the state-of-the-art methods.

\end{itemize}

\section{Related Work}

The objective of the empathetic response generation is to equip the model with the capability to deliver suitable emotional value\cite{hofmann2010empirical}. Numerous work has addressed this task in terms of perception enhancement and cognition enhancement. 

\subsection{Perception Enhancement Methods}
Methods for perception enhancement typically analyze the sentiment words in conversations and distill fine-grained emotion control information\cite{lin2019moel, ghosal2020cosmic, inproceedings, zhao2022don}.
For example, multi-granularity sentiment labels of conversation and utterance in data can help the model capture user sentiment efficiently\cite{wang2022empathetic}. 
E-CORE\cite{fu2023core} focus on exploring intrinsic sentiment by emotion correlation learning, utilization, and supervising. Besides, CAB\cite{gao2023cab} split the empathy response generation into three parts: cognition, affection and behavior. Additionally, ESCM\cite{yang-etal-2023-exploiting-emotion} uses dynamic emotion-semantic vectors and dependency trees to guide the model generate empathetic responses. Furthermore, \citet{yufeng2024ctsm} categorizing emotions into fine-grained trait and state emotion to control sentiment representation more accurately. 
Nevertheless, despite the potential for fine-grained emotional control data to reveal perspective-specific emotion states, the sensible representation of dialogue at the utterance level remains underutilised.  

Based on the self-presentation theory in psychology, we divide the conversation into sentences with two types of attributes: sensible and rational, and explore their importance for empathetic response\cite{fritz1998distinctions, smith2006cognitive}.

\subsection{Cognition Enhancement Methods}

Cognitive enhancement can improve the model's comprehension and learning capacity.\cite{tahir2023artificial, zhou2022case}. For example, CEM \cite{sabour2022cem} introduces commonsense knowledge inference to this task for the first time through the pre-trained COMET model. 
The scope of cognition enhancement methods is further expanded with the help of ambiguous information filtering\cite{cai2023improving} and graph neural network structure\cite{li2022knowledge}. However, these approaches are constrained by limitations of semantic contraction and single-role  view\cite{cai2023improving}. 

Chain-of-Thought Prompting(CoT) are typically used to facilitate the logical resolution of complex problems, elucidating the rationale behind the decision-making process\cite{wei2022chain, wang2022self, wang-etal-2023-towards}. The chain-of-empathy method employs CoT technology to enhance the cognitive capacity of the empathetic response model, though it does not examine the logical interrelationship between emotion and conversation history\cite{lee2023chain}.
In our work, we utilize the chain of emotion-aware empathetic prompting method to enhance the cognitive ability of the model.

\section{Method}

\begin{figure*}
\centering 
\includegraphics[width=1.0\textwidth]{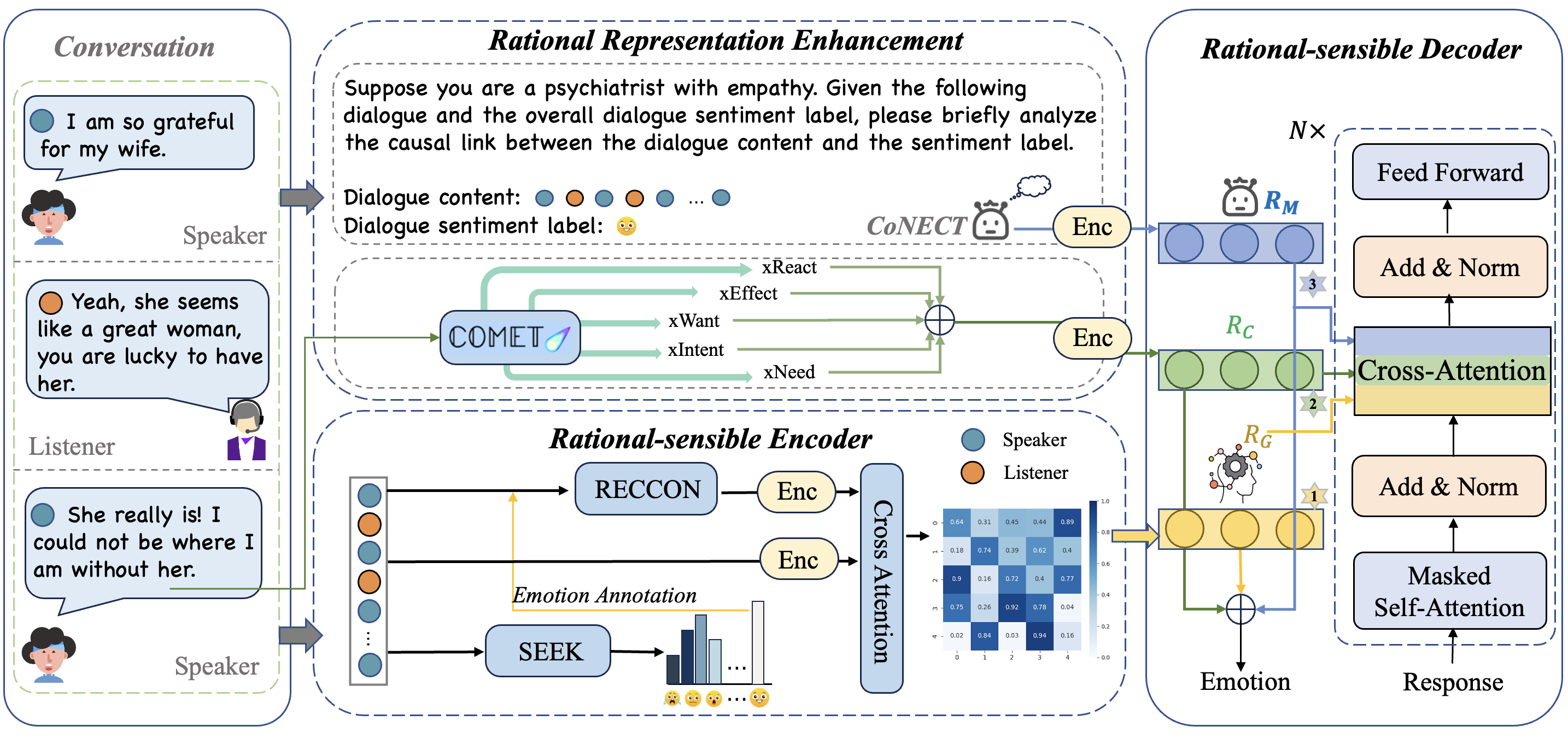} 
\caption{The model architecture of Lamb, which consists of three parts: (a) Rational-sensible Encoder based on self-presentation theory, (b) Rational Representation Enhancement by CoNECT and COMET, (c) Rational-sensible Decoder.}
\label{model_structure}
\end{figure*}

As shown in Figure~\ref{model_structure}, Lamb is consists of three models: 1) rational-sensible encoder, 2) rational representation enhancement, 3) rational-sensible decoder.
\subsection{Task Formulation}
Given a dialogue history $U = [S_1, L_1, S_2, L_2, ..., L_{N-1}, S_N]$ of 2$N$-1 utterances, the goal of the empathetic response generation is to predict the overall sentiment label $e_{tar}$ and generate empathetic response $Y = [y_1, y_2, ..., y_{m_y}]$. In dialogue history $U$, $S_i = [s_1^i, s_2^i, ..., s_{m_{s_i}}^i]$ and $L_i = [l_1^i, l_2^i, ..., l_{m_{l_i}}^i]$ represent the $i-$th utterance of speaker and listener, composed by $m_{s_i}$ and $m_{l_i}$ tokens respectively.

\subsection{Rational-sensible Encoder}


The first group encompasses sentences that used to underscore individual characteristics by sensible expression. On the other hand, the second group is composed of sentences that primarily emphasize the rational reasoning. When selecting the first group sentences, our purpose is to align the global sensible experiences of both the speaker and listener, which enables the model to better understand the speaker's emotion and respond more empathy. Thereforce, we adopt the pretrained model SEEK\citep{wang2022empathetic} to prelabel comprehensive sentiment for each conversation, inspired by  \citet{welivita2020taxonomy}. Then, we use a pretrained emotion-cause model RECCON\citep{poria2021recognizing} to detect relevant sensible expression in context.

As described in Figure~\ref{model_structure}, we set $U$ as input of the SEEK model, and obtain predicted global sentiment label $e_{ano}$ accordingly. Next, we use $e_{ano}$ as the target sentiment of the model RECCON, and look for cues that evoke the sentiment in $U$. The final output is the sensible sentences $D$ that implies the expression of the sentiment, which is a subset of $U$.

\begin{equation}
    e_{ano} = SEEK(U)
\end{equation}

\begin{equation}
    D = RECCON(e_{ano}, U)
\end{equation}

We feed $U$ and $D$ into the Bart encoder to obtain the context representation $R_U \in \Re^{l_U\times d}$ and the cause representation $R_D \in \Re^{l_D\times d}$:

\begin{equation}
    R_U = Bart_{enc}(S_1 \oplus \langle \verb|\| s \rangle \oplus L_1 \oplus \langle \verb|\| s \rangle ... S_N)
\end{equation}

\begin{equation}
    R_D = Bart_{enc}(D)
\end{equation}

Additionally, in order to assist the model in discerning between these two categories of sentences, we incorporate the method of joint modeling through the designed attention mechanism to acquire distinct weights for them, and output finally representation $R_G \in \Re^{l_G\times d}$.

\begin{equation}
    R_G = softmax(\frac{(W_q R_U)\cdot (W_k R_D)^T}{\sqrt{2d}})\cdot(W_v R_D)
\end{equation}
, where $W_q \in \Re^{d\times d}$, $W_k \in \Re^{d\times d}$ and $W_v \in \Re^{d\times d}$ are randomly initialised tensor matrix. $l_U$, $l_D$ and $l_G$ is the number of corresponding tokens. $d$ is the dimension of feature vector.

\subsection{Rational Representation Enhancement}
In this module, we dedicate to enhancing rational information by CoNECT and COMET's commonsense knowledge. 

\subsubsection{COMET}
Following previous works, we adopt the pretrained model COMET to generate common sense knowledge. 
Given the last utterance $S_N$ of the historical context $U$, we alternately select xIntent, xEffect, xWant, xReact and xNeed cognitive attributes for inferencing, obtaining $C_r$ commonsense knowledge where $r \in \{xIntent, xEffect, xWant, xReact, xNeed\}$:
\begin{equation}
C_r = COMET(S_N)
\end{equation}
After that, we append the $[CLS]$ token to the first position for each COMET relation. Then we fed them into the encoder getting feature matrix $R_C \in \Re^{l_C\times d}$, where $l_C$ is the number of tokens in set $r$ and special tokens.

\begin{equation}
    R_C = Bart_{enc-r}([CLS] \oplus C_r)
\end{equation}

As a result, we embed the commonsense knowledge in the decoder, so as to provide more effective complementary information to the model.

\subsubsection{CoNECT}


As shown in Figure~\ref{model_structure}, the CoT prompt we used can be divided into three parts:

\textbf{Character.}  We assign LLM to role-play a compassionate psychologist who is expected to possess theoretical knowledge and behavioral paradigms in professional domain.  

\textbf{Causal Chain.}   In order to improve contextual coherence and achieve more precise comprehension of dialogues, we extend the scope of external knowledge sources by incorporating the complete context into the prompt.

\textbf{Global Sentiment Label.} In conjunction with the causal chain, the method offers guidance for the logical behavior of the analytical model.

In practice, we use the prompt $C_P$ as the input of LLM, and the output of the LLM is the CoNECT data $C_M$.

\begin{equation}
    C_P = Prompt_{template}(U, e_{ano})
\end{equation}

\begin{equation}
    C_M = LLM(C_P)
\end{equation}

Afterwards, we pass $C_M$ through the encoder to acquire unified feature representation $R_M \in \Re^{c\times d}$, where $c$ is the length of CoNECT data.

\begin{equation}
    R_M = Bart_{enc}(C_M)
\end{equation}

\subsection{Rational-sensible Decoder}

The rational-sensible decoder facilitates the model's ability to balance sensible and rational cognition, thereby generating empathetic responses that are more consistent with the listener's role. 



To begin with, the model perceives the sensible and rational cognitive aspects of the context with $R_G$, which offers essential background information. Subsequently, the model extends the basic commonsense knowledge with the assistance of the $R_C$. Lastly, the model incorporates information from the $R_M$ to enrich and intensify the rational cognitive expression.

The target response $Y=[y_1, y_2, ..., y_{m_y}]$ with length $m_y$, which is generated token by token by Bart decoder based above representation $R_G$, $R_C$ and $R_M$. During training, we adopt the standard negative log-likelihood(NLL) loss on the target response $Y$:
\begin{equation}
    \pounds_{nll} = - \sum_{t=1}^{m_y}\log(y|(R_G, R_C, R_M), y_{<t}) 
\end{equation}

\subsection{Training Objectives}

The objective of our training program is composed of two parts: emotion classification and empathetic response generation.

\subsubsection{Emotion Classification}

For emotion classification task, we utilize average pooling to get the knowledge vector $p_k \in \Re^{d}$:

\begin{equation}
    p_k = Average-pooling(R_C)
\end{equation}

Then, in order to optimize knowledge acquisition, we strive for the fusion of historical conversations, empathetic thought chains, and common sense knowledge.

\begin{equation}
    R_F = R_G[0]\oplus R_M[0]\oplus p_k,
\end{equation}

\begin{equation}
    P_{e_{pre}} = softmax(W_{\theta}(R_F))
\end{equation}
After that, we subsequently pass $R_F$ through a linear operation $W_\theta \in \Re^{d\times q}$, followed by a softmax calculation to produce the distribution $P_{e_{pre}} \in \Re^q$, where $q$ is the number of emotion categories:

In training phase, we optimize the object by minimizing the Cross-Entropy loss between the emotion category and the ground truth label $e_{tar}$:

\begin{equation}
    \pounds_{emo} = -log(P_{e_{pre}}(e_{tar}))
\end{equation}

\subsubsection{Final Training Objective}

In the training phase, all the parameters of our model are optimized with $\pounds_{emo}$ and $\pounds_{nll}$:

\begin{equation}
    \pounds = \pounds_{nll} + \pounds_{emo}
\end{equation}

\section{Experiments}

\begin{table*}
  \centering
  \caption{Results of automatic evaluation. Optimal outcomes are highlighted in bold, and suboptimal outcomes are indicated by underlining. The evaluation of the LLM utilizes a meticulously crafted prompt for inference.}
  \resizebox{\linewidth}{!}{
    \begin{tabular}{ccccccccccc}
    \toprule
    \textbf{Models} & \textbf{PPL} & \textbf{B-1} & \textbf{B-2} & \textbf{B-3} & \textbf{B-4} & \textbf{R-1} & \textbf{R-2} & \textbf{Dist-1} & \textbf{Dist-2} & \textbf{Acc} \\
    \midrule
    MoEL\citep{lin2019moel}  & 36.60 & 18.07 & 8.30  & 4.37  & 2.65  & 18.24 & 4.81  & 0.59  & 2.64  & 31.74 \\
    MIME\citep{ghosal2020cosmic}  & 37.24 & 18.60 & 8.39  & 4.54  & 2.81  & 17.08 & 4.05  & 0.47  & 1.66  & 30.96 \\
    EmpDG\citep{inproceedings} & 37.43 & 19.96 & 9.11  & 4.74  & 2.80  & 18.02 & 4.43  & 0.46  & 1.99  & 31.65 \\
    CEM\citep{sabour2022cem}   & 36.33 & 16.12 & 7.29  & 4.06  & 2.03  & 15.77 & 4.50  & 0.62  & 2.39  & 36.84 \\
    SEEK\citep{wang2022empathetic}  & 36.78 & 10.77 & 4.40  & 2.02  & 1.08  & 12.74 & 2.94  & 0.68  & 2.81  & 42.74 \\
    CASE\citep{zhou2022case}  & 35.20 & 15.59 & 7.22  & 3.80  & 2.24  & 17.33 & 4.67  & 0.65  & 3.37  & 38.99 \\
    E-CORE\citep{fu2023core} & 33.03 & -   & -   & -   & -   & -   & -   & 0.72  & 3.49  & 42.59 \\
    KEMP\citep{li2022knowledge}  & 36.39 & 16.72 & 7.17  & 3.77  & 2.33  & 16.11 & 3.31  & 0.66  & 3.07  & 36.57 \\
    CAB\citep{gao2023cab}   & 34.36 & \textbf{19.23} & 8.55  & 4.36  & 2.57  & 17.50 & 4.13  & 1.13  & 4.23  & 40.52 \\
    ESCM\citep{yang-etal-2023-exploiting-emotion}  & 34.82 & -   & -   & -   & -   & -   & -   & 1.19  & 4.11  & 41.19 \\
    DCKS\citep{cai2023improving}  & \textbf{18.58} & 18.75 & \textbf{9.12}  & \textbf{5.38}  & \textbf{3.57}  & \textbf{19.14} & \textbf{5.45}  & 1.57  & 6.02  & \textbf{48.69} \\
    CTSM\citep{yufeng2024ctsm}  & 34.56 & -   & -   & -   & -   & -   & -   & \textbf{2.00}  & \textbf{7.34}  & 43.41 \\
    \midrule
    Qwen1.5-MoE-A2.7B-Chat\cite{qwen1.5} & -   & 7.73  & 2.39  & 1.03  & 0.52  & 10.45 & 1.12  & 2.76  & 22.41 & - \\
    LLaMA2-13B-Instruct\citep{touvron2023llama} & -   & 11.69 & 4.03  & 1.79  & 0.93  & 13.27 & 1.83  & 2.91  & 18.92 & - \\
    LLaMA3-8B-Instruct\citep{{touvron2023llama}} & -   & 13.17 & 4.42  & 1.92  & 1.02  & 14.12 & 1.68  & 2.69  & 18.70 & - \\
    Qwen1.5-72B-Chat\cite{qwen1.5} & -   & 14.19 & 4.85  & \textbf{2.27}  & \textbf{1.23}  & 13.83 & \textbf{1.97}  & 3.29  & 22.68 & - \\
    Mixtral-8x7B-Chat\citep{jiang2024mixtral} & -   & \textbf{14.66} & \textbf{4.72}  & 2.11  & 1.10  & \textbf{14.54} & 1.69  & 3.30  & 21.13 & - \\
    \midrule
    \textbf{Lamb(CoNECT based on LLaMA2-13B-Instruct)} & \underline{20.14}  & \boxed{\textbf{22.16}} & \boxed{\textbf{10.55}} & \boxed{\textbf{6.01}} & \boxed{\textbf{3.78}} & \boxed{\textbf{19.39}} & \boxed{\textbf{5.47}} & \boxed{\textbf{2.38}} & \boxed{\textbf{10.32}} & \boxed{\textbf{51.36}} \\
    \rowcolor{cyan!15} 
    \textit{+ Compared with LM} & -   & $\blacktriangle$ 2.93 & $\blacktriangle$ 1.43  & $\blacktriangle$ 0.63  & $\blacktriangle$ 0.21  & $\blacktriangle$ 0.25 & $\blacktriangle$ 0.02  & $\blacktriangle$ 0.38  & $\blacktriangle$ 2.98 & $\blacktriangle$ 2.67 \\
    \rowcolor{cyan!15} 
    \textit{+ Compared with LLM} & -   & $\blacktriangle$ 7.50 & $\blacktriangle$ 5.83  & $\blacktriangle$ 3.74  & $\blacktriangle$ 2.55  & $\blacktriangle$ 4.85 & $\blacktriangle$ 3.50  & -  & - & - \\

    \midrule
    \textbf{\& CoNECT based on LLaMA3-8B-Instruct} & \underline{20.22}  & \boxed{\textbf{22.56}} & \boxed{\textbf{11.03}} & \boxed{\textbf{6.33}} & \boxed{\textbf{3.82}} & \boxed{\textbf{19.73}} & \boxed{\textbf{5.68}} & \boxed{\textbf{2.41}} & \boxed{\textbf{10.41}} & \boxed{\textbf{52.16}} \\
    \rowcolor{cyan!15} 
    \textit{+ Compared with LM} & -   & $\blacktriangle$ 3.32 & $\blacktriangle$ 1.91  & $\blacktriangle$ 0.95  & $\blacktriangle$ 0.25  & $\blacktriangle$ 0.59 & $\blacktriangle$ 0.23  & $\blacktriangle$ 0.41  & $\blacktriangle$ 3.07 & $\blacktriangle$ 3.47 \\
    \rowcolor{cyan!15} 
    \textit{+ Compared with LLM} & -   & $\blacktriangle$ 7.87 & $\blacktriangle$ 6.31  & $\blacktriangle$ 4.06  & $\blacktriangle$ 2.59  & $\blacktriangle$ 5.19 & $\blacktriangle$ 3.71  & -  & - & - \\

    \midrule
    \textbf{\& CoNECT based on Qwen1.5-72B-Chat} & \underline{20.15}  & \boxed{\textbf{22.27}} & \boxed{\textbf{10.88}} & \boxed{\textbf{6.32}} & \boxed{\textbf{3.60}} & \boxed{\textbf{19.54}} & \boxed{\textbf{5.71}} & \boxed{\textbf{2.29}} & \boxed{\textbf{10.33}} & \boxed{\textbf{51.69}} \\
    \rowcolor{cyan!15} 
    \textit{+ Compared with LM} & -   & $\blacktriangle$ 3.04 & $\blacktriangle$ 1.76  & $\blacktriangle$ 0.94  & $\blacktriangle$ 0.03  & $\blacktriangle$ 0.40 & $\blacktriangle$ 0.26  & $\blacktriangle$ 0.29  & $\blacktriangle$ 2.99 & $\blacktriangle$ 3.0 \\
    \rowcolor{cyan!15} 
    \textit{+ Compared with LLM} & -   & $\blacktriangle$ 7.61 & $\blacktriangle$ 6.16  & $\blacktriangle$ 4.05  & $\blacktriangle$ 2.37  & $\blacktriangle$ 5.0 & $\blacktriangle$ 3.74  & -  & - & - \\

    \midrule
    \textbf{\& CoNECT based on Mixtral-8x7B-Chat} & \underline{20.18}  & \boxed{\textbf{22.24}} & \boxed{\textbf{10.87}} & \boxed{\textbf{6.35}} & \boxed{\textbf{3.79}} & \boxed{\textbf{19.51}} & \boxed{\textbf{5.73}} & \boxed{\textbf{2.37}} & \boxed{\textbf{10.35}} & \boxed{\textbf{51.75}} \\
    \rowcolor{cyan!15} 
    \textit{+ Compared with LM} & -   & $\blacktriangle$ 3.01 & $\blacktriangle$ 1.75  & $\blacktriangle$ 0.97  & $\blacktriangle$ 0.22  & $\blacktriangle$ 0.37 & $\blacktriangle$ 0.28  & $\blacktriangle$ 0.37  & $\blacktriangle$ 3.01 & $\blacktriangle$ 3.06 \\
    \rowcolor{cyan!15} 
    \textit{+ Compared with LLM} & -   & $\blacktriangle$ 7.58 & $\blacktriangle$ 6.15  & $\blacktriangle$ 4.08  & $\blacktriangle$ 2.56  & $\blacktriangle$ 4.97 & $\blacktriangle$ 3.76  & -  & - & - \\
    
    \bottomrule
    \end{tabular}%
    }
  \label{tab:main_exp}%
\end{table*}%

\subsection{Datasets}


Our experiments are conducted on the widely used EMPATHETICDIALOGUES\citep{rashkin2018towards}, comprising 25k multi-turn empathetic conversations between a speaker and a listener, with an average of 4.31 turn per dialog. This dataset provides 32 evenly distributed emotion labels, and each conversation is assigned a related label.

To ensure the fairness of the comparison experiments, we use the same dataset division as the previous research method, dividing the dataset into training, validation and test sets in an 8:1:1 ratio\citep{cai2023improving}.

\subsection{Evaluation Metrics}
In order to confirm the effectiveness of Lamb, we implement a dual evaluation strategy, a comprehensive evaluation from  automation and human perspectives. 
The detailed experiment setting is described in Appendix~\ref{sec:compared method}.
And the compared baseline models are listed in Appendix~\ref{sec:baseline methods}. 

\subsubsection{Automatic Evaluation}
Following previous work, we choose Perplexity(\textbf{PPL}), corpus-level BLEU(\textbf{B-N}), sentence-level ROUGE(\textbf{R-N}), Distinct-n(\textbf{Dist-N}) and Accuracy(\textbf{Acc}) as our main automatic metrics. 1) Perplexity is used to assess the fluency and comprehensibility of the text generated by the model, and a lower PPL value means that the text is more natural. 2) The BLEU and ROUGE score indicate the degree of similarity between the generated text and the ground-truth text. A higher score indicates a greater degree of similarity. 3) Dist-N is used to assess the diversity of the content, with higher metric values indicating that the model can produce more diverse and richer representation. 4) Acc is used to measure the accuracy of the model for emotion classification.


\subsubsection{Human Evaluation}


To further verify the effectiveness and reliability of our method, we conduct manual evaluation work. Specially, we set four evaluation dimensions, and the score of each dimension is quantitatively evaluated from 1 to 5. 
1) Coherence(\textbf{Coh.}): Measures the level of relevance between the text generated by the model and the gold response.
2) Empathy(\textbf{Emp.}): Demonstrates a greater comprehension of the speaker's circumstances and conveys a more fitting sentiment.
3) Informativeness(\textbf{Inf.}): Measures the richness of information contained in generated responses.
4) Continuity(\textbf{Cont.}): Indicates the intensity of the speaker's desire to continue the conversation.
Particularly, we randomly select 1000 sets of test data and randomly shuffled their order to ensure the fairness and objectivity of the evaluation. Then we distribute the data to three qualified and experienced human evaluators, whom score the model's responses and get a average score. In this way, we can ensure the accuracy and scientificity of the assessment results, which in turn strongly supports the effectiveness of our methods. The detailed basis of scoring is shown in the Appendix~\ref{sec:human eva}.



\subsection{Automatic Evaluation Results}

\begin{table*}
  \centering
  \caption{Ablation study}
  \resizebox{\linewidth}{!}{
    \begin{tabular}{lcccccccccc}
    \toprule
    \textbf{Models} & \textbf{PPL $\downarrow$} & \textbf{B-1 $\uparrow$} & \textbf{B-2 $\uparrow$} & \textbf{B-3 $\uparrow$} & \textbf{B-4 $\uparrow$} & \textbf{R-1 $\uparrow$} & \textbf{R-2 $\uparrow$} & \textbf{Dist-1 $\uparrow$} & \textbf{Dist-2 $\uparrow$} & \textbf{Acc $\uparrow$} \\
    \midrule
    Vanilla  & 18.58 & 18.75 & 9.12  & 5.38  & 3.57  & 19.14 & 5.45  & 1.57  & 6.02  & 48.69 \\
    Vanilla + Self-pres & 17.74 & 21.02 & 10.17 & 5.84  & 3.74  & 19.88 & 5.56  & 2.01  & 8.52  & 50.65 \\
    Vanilla + CoNECT & 20.30 & 19.95 & 9.79  & 5.72  & 3.69  & 19.62 & 5.69  & 2.25  & 9.96  & 51.30 \\
    Lamb  & 20.14 & 22.16 & 10.55 & 6.01  & 3.78  & 19.39 & 5.22  & 2.38  & 10.32 & 51.36 \\
    \bottomrule
    \end{tabular}%
    }
  \label{tab:ablation}%
\end{table*}%

\begin{figure*}
\centering 
\includegraphics[width=1\textwidth]{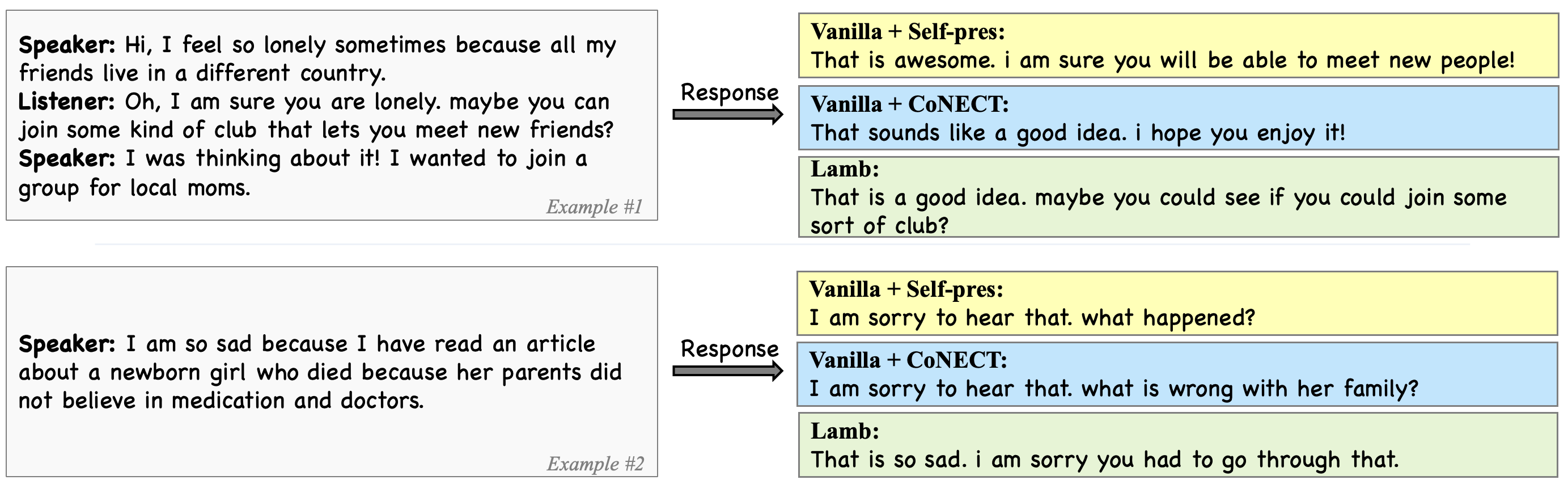} 
\caption{Effect of Self-presentation and CoNECT module.}
\label{Fig.abalation_case_img}
\end{figure*}

Our CoNECT module conduct several sets of experiments based on different LLMs: LLaMA2-13B-Instruct, LLaMA3-8B-Instruct, Qwen1.5-72B-Chat and Mixtral-8x7B-Chat. Compared with LM: 
MoEL,
 MIME,
 EmpDG,
 CEM,
 SEEK,
 CASE,
 E-CORE,
 KEMP,
 CAB,
 ESCM,
 DCKS, and
 CTSM,
 Lamb exceed the baseline models on most metrics, proving the robustness and effectiveness of our method. We observe that all four variants of Lamb achieve significant performance, which indicates that the improvement comes from the architecture of Lamb rather than relying on a specific LLM.

The enhancement in both Dist-1 and Dist-2 indicates a greater diversity in our model responses, thereby mitigating the appearance of hollow response. Furthermore, achieving the highest scores on both BLEU-N and ROUGE-N metrics reflects better consistency between model responses and historical conversations while also maintaining diversity. Additionally, our model exhibits a notable increase of 2.67 points in emotion classification accuracy, showcasing its capability to accurately capture speaker sentiment. Despite a slight decrease in our PPL due to the complex structure of the Lamb, the overall improvement in other evaluation metrics reaffirms the effectiveness of our approach. We will discuss the results of the comparison with LLM in Section~\ref{comparing with llm}.

\subsection{Ablation Studies}

In order to prove the validity of the CoNECT and self-presentation module, we conduct two ablation experiments. The following is an introduction to the comparison models used in the experiment:

1) \textbf{Vanilla}: Removing both Self-presentation module and CoNECT module.

2) \textbf{Vanilla + Self-pres}: Based on vanilla, the self-presentation module is added.

3) \textbf{Vanilla + CoNECT}: Based on vanilla, the CoNECT module is added.



The Table~\ref{tab:ablation} demonstrates that removing either the Self-presentation module or the CoNECT module result in a significant decrease in most evaluation metrics, indicating the effectiveness of both modules. When comparing the CoNECT and Self-presentation module, 
the approach with CoNECT excels in diversity metrics but underperforms in BLUE and ROUGE metrics.
This observation suggests that the CoNECT module enhances the model by providing more contextual information with external knowledge. The Self-pres module, on the other hand, directs the model to focus more on the dialog content to improve semantic understanding. The synergistic effect of these two modules boosts the model's performance in empathetic response generation.

\begin{figure}
\centering 
\includegraphics[width=0.47\textwidth]{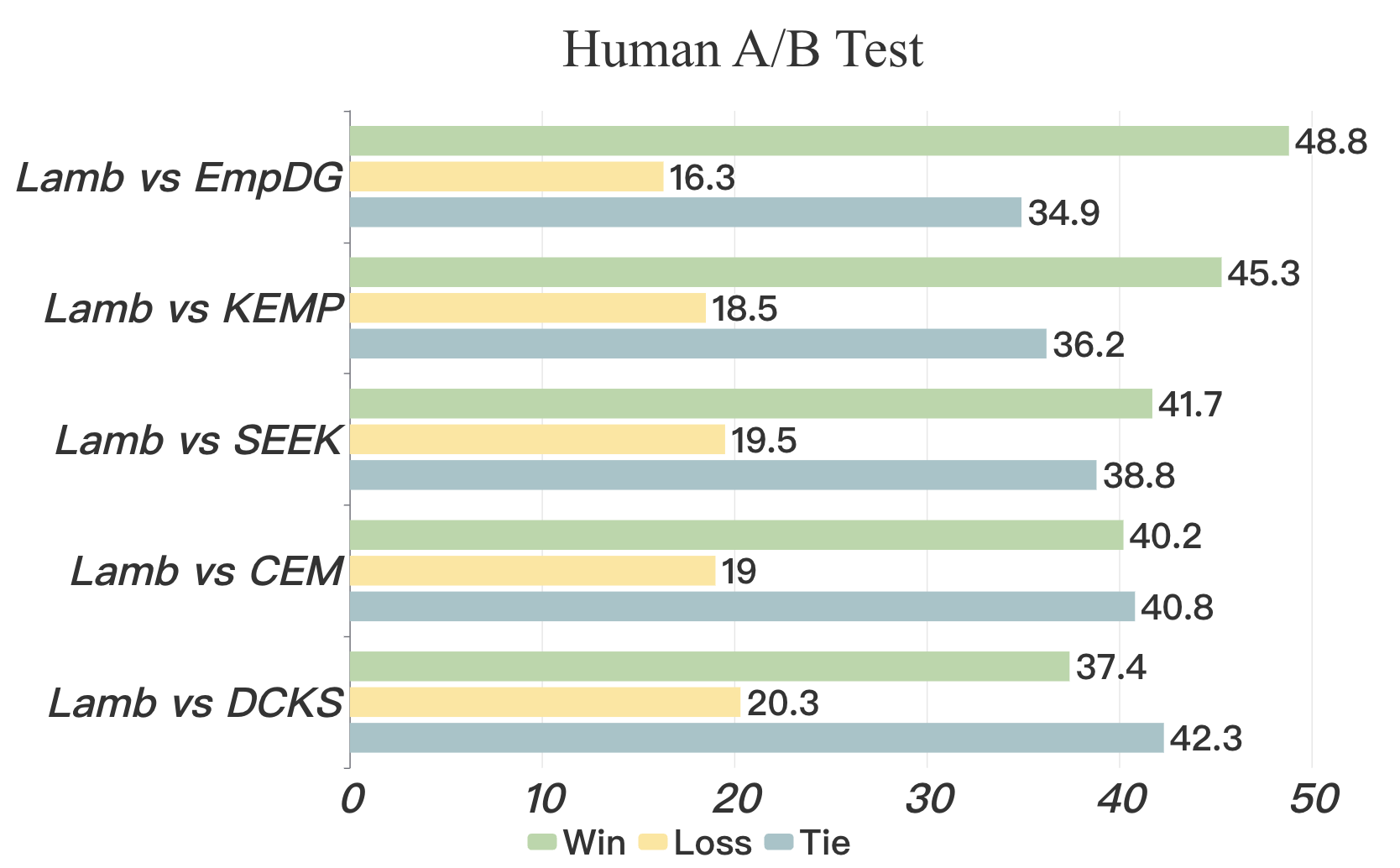} 
\caption{Results of Human A/B test.}
\label{Fig.human_ab_fig}
\end{figure}



To further validate the effectiveness of our proposed self-pres and CoNECT modules, we provide a detailed analysis with two sets of case ablation experiments, as shown in Figure~\ref{Fig.abalation_case_img}.

\textbf{Self-presentation Module. }
In example-1, the model integrates both perceptual and rational cognition by conducting a detailed analysis of the conversation context to better understand and respond to the user's expressed emotions of "longing for friends".   This approach not only strengthens the emotional bond between the model and the user but also proves the effectiveness of the self-presentation theory in real-world scenarios.  
In example-2, the model effectively utilizes self-presentation module to express empathy, and leads speaker to express their deeper feeling of sadness by asking questions. This strategy not only demonstrates the model's role as a superb listener but also emphasizes its human-like qualities.   

\textbf{CoNECT. }
In example-1, CoNECT recognizes that users were joining social groups out of loneliness. The model responds positively to the decision, expressing optimism about its potential to enhance user satisfaction. These responses not only address the emotional needs of users but also demonstrate CoNECT's expertise in handling complex emotional issues.  
In example-2, CoNECT analyzes parents' distrustful attitude towards medication and doctor's advice, clearly pointing out that this is a key factor leading to the unfortunate death of a newborn.   As a result, the model responds by  questioning family members to provide emotional support to the user.  

\begin{table}[htbp]
  \centering
  \caption{Results of human evaluation.}
    \resizebox{\linewidth}{!}{
    \begin{tabular}{l|cccc}
    \toprule
    \textbf{Models} & \textbf{Coh.} & \textbf{Emp.} & \textbf{Inf.} & \textbf{Cont.} \\
    \midrule
    EmpDG &	3.22  &	3.10  &	2.99  &	3.07\\
    CEM	  & 3.41  &	3.49  &	3.12  &	3.18 \\
    SEEK  &	3.40  &	3.62  &	3.19  &	3.33 \\
    KEMP  &	3.56  &	3.66  &	3.35  &	3.52 \\
    DCKS  &	4.10  &	3.94  &	3.76  &	3.87 \\
    \midrule
    \textbf{Lamb} & \textbf{4.31} & \textbf{4.21} & \textbf{4.78} & \textbf{4.32} \\
    \rowcolor{cyan!15} 
    + \textit{Improvement} & + 0.21 & + 0.27 & + 1.02 & + 0.45 \\
    \bottomrule
    \end{tabular}%
    }
  \label{tab:huamn_exp}%
\end{table}%

\subsection{Human Evaluation Results}
 

As illustrated in Table~\ref{tab:huamn_exp}, Our model improves on all human assessment indicators. The incorporation of CoNECT enhancements  the model's capacity for empathizing, along with an increase in the volume of information conveyed in the response. Furthermore, reinforced by self-presentation theory, the model can focus on the sensibility clues in historical content, improving the contextual coherence of the generated response and thus motivating the speaker's willingness to continue the conversation. Additionally, the pairwise response comparison results are shown in Figure~\ref{Fig.human_ab_fig}. The results further confirm that the responses generated by Lamb are more preferred by human judges. Please refer to the Appendix~\ref{sec:Case Study} for a more detailed case study.

\begin{figure}
\centering 
\includegraphics[width=0.49\textwidth]{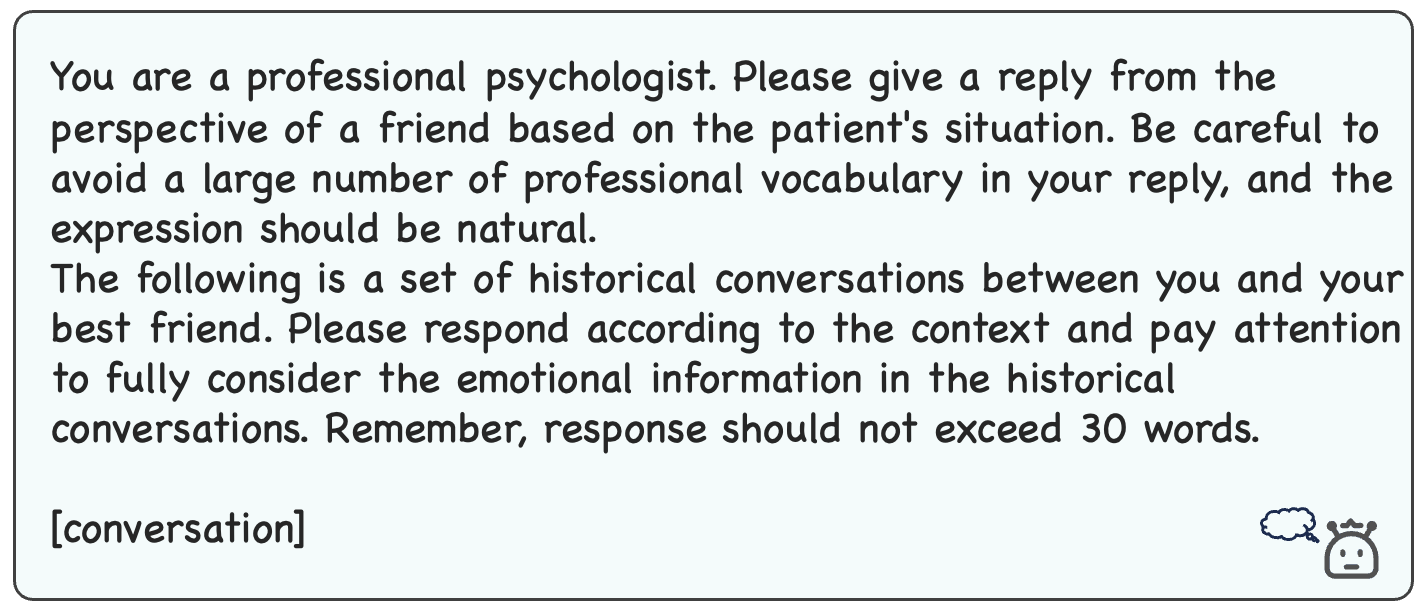} 
\caption{Prompt used for evaluating LLM empathetic ability.}
\label{Fig.eval_prompt}
\end{figure}

\begin{table}[htbp]
  \centering
  \caption{Compared with closed source LLMs.}
  \resizebox{0.88\linewidth}{!}{
    \begin{tabular}{l|cc}
    \toprule
    \textbf{Models} & \textbf{BLEU-2} & \textbf{BLEU-4} \\
    \midrule
    ChatGPT(+ 0-short)  & 6.19  & 1.86 \\
    ChatGPT(+ 1-short)  & 6.79  & 2.12 \\
    ChatGPT(+ 5-short)  & 7.85  & 2.65 \\
    GPT-3(+ 0-short)  & 6.88  & 2.22 \\
    GPT-3(+ 1-short)  & 6.71  & 2.16 \\
    GPT-3(+ 5-short)  & 8.51  & 2.83 \\
    GPT-3.5(+ 0-short)  & 8.51  & 2.80 \\
    GPT-3.5(+ 1-short)  & 5.62  & 1.99 \\
    GPT-3.5(+ 5-short)  & 9.37  & 3.26 \\
    GPT-4(+ 0-short)  & 9.83  & 3.25 \\
    \textbf{Lamb}  & \textbf{10.55} & \textbf{3.78} \\
    \bottomrule
    \end{tabular}%
    }
  \label{tab:main_llm}%
\end{table}%

\subsection{Comparing with LLM}
\label{comparing with llm}
Although we employ LLaMA2-13B-Chat inference data as an aid during our study, our generative model is still based on BART, a much less parameters model than LLM.
Considering the large amount of data processed during LLM training and the sampling strategy used, there is a significant advantage over small models, thus, we do not compare perplexity and diversity metrics with them.

The compared open-source LLMs utilize a meticulously crafted system prompt in order to optimising its capacity of empathetic response, as shown in Figure~\ref{Fig.eval_prompt}.
In the bottom of Table~\ref{tab:main_exp}, our model is best in all BLUE and ROUGE metrics.
Furthermore, our model outperforms compared  closed-source LLM in BLUE-2 and BLUE-4 in Table~\ref{tab:main_llm}. For fairness, we maintain the setting of GPT-series LLM are consistent of \cite{qian2023harnessing}.




\section{Conclusions}
In this research, we enhance the generation of empathetic responses through the incorporation of self-presentation module and CoNECT. The results of both automated and human evaluations indicate the superior quality of responses generated by Lamb. 

In the future, we will strive to strengthen the integration of multidisciplinary knowledge and further delve the significance of sensibility and rationality in cognitive processes. 

\section{Limitations}
There are two main limitations in our work. First of all, we employed COMET commonsense knowledge in the rational representation enhancement module, however, we did not take a strong approach to mitigate the problem of semantic conflicts between the information it generates. Secondly, automated evaluation is difficult to score the empathy of generated responses, existing a certain degree of limitation.
\section{Ethics Considerations}
We obtain our data from EMPATHETICDIALOGUES, a publicly available dataset that does not include any personal information.   Our human assessments are carried out by three experienced annotators and they are fairly compensated for their work.






\clearpage
\bibstyle{acl_natbib} 
\bibliography{anthology}

\clearpage

\newpage
\appendix
\section{Case Study}
\label{sec:Case Study}
The responses generated by Lamb and other models are shown in Table~\ref{tab:main_case}. The response generated by KEMP is not relevant to the history of the dialog. Although SEEK and DCKS responses expressed congratulations to the speaker, they did not establish a connection to concrete things in the conversation and therefore contained relatively few semantic information.

In our method, CoNECT analyzes the speaker's state of mind from a global perspective, and speculates that "the happiest thing in life" and "something you will never forget" are the key words in the conversation, which add color to the joy brought by the birth of the first child. The keywords emphasize the joy of the birth of the first child. By further integrating the self-presentation theory, Lamb recognized the speaker's joyfulness while closely focusing on the key messages in the conversation, which effectively stimulated the speaker's willingness to continue the communication. Table~\ref{tab:appendix_case1} and Table~\ref{tab:appendix_case2} are two more empathetic response case.

In comparison to responses from LLaMA2-13B-Chat that demonstrate strong emotion, our response has the following advantages:

(1) Lamb directly expresses sympathy for Speaker's feelings. This simplicity avoids excessive explanation or unnecessary emotional burden that might be caused.

(2) In Table~\ref{tab:appendix_case1}, Lamb echoes Speaker's feelings about “finally meeting someone after waiting for a long time” without shifting to Listener's own experience. This allows the Speaker to feel that their feelings have been responded to and valued.

(3) The information presented in Table~\ref{tab:main_case} shows that LLaMA2-13B-Chat shares personal experiences. This can make the speaker feel like the other person is talking about themselves instead of paying attention to the speaker's feelings. As a result, this may divert the conversation away from the original emotional expression of the speaker.

\section{Baseline Models Details}
\label{sec:baseline methods}
The following are the models we compared in the experiments. We use the official codes and follow the implementations.

1) MoEL\citep{lin2019moel}: The model designs a corresponding decoder for each emotion to generate a response, and all the results are synthesised to produce the final result.
2) MIME\citep{ghosal2020cosmic}: Depending on the positive or negative polarity of the emotion and the contextual content, the model is able to simulate the user's emotions and enable the generation of empathetic responses by introducing randomness.
3) EmpDG\citep{inproceedings}: A model that includes an empathetic information generator and a sentiment discriminator. The function of the generator is to ensure that the model has the ability to generate diverse content, while the function of the discriminator is to ensure that the information produced by the generator matches the empathetic sentiment in the context.
4) CEM\citep{sabour2022cem}: The first model introduces the COMET pre-trained model to acquire common sense knowledge in empathetic response generation task, and uses both intent and cognitive types to categorise the knowledge.
5) SEEK\citep{wang2022empathetic}: A model focuses on sentence-level sentiment information and utilizes attention mechanisms to provide the model with multi-granularity sentiment label information.
6) CASE\citep{zhou2022case}: A model employ external message, COEMT and ConceptNet, to enhance the ability of cognitive and emotion.
7) E-CORE\citep{fu2023core}: A method fouces on exploring intrinsic sentiment by emotion correlation learning, utilization, and supervising. 
8) KEMP\citep{li2022knowledge}: A model that uses ConceptNet and VRC-NED as external knowledge sources and performs contextual modelling through a graph neural network structure.
9) CAB\citep{gao2023cab}: A model split the empathy response generation into three parts: cognition, affection and behavior. 
10) ESCM\citep{yang-etal-2023-exploiting-emotion}: Using dynamic emotion-semantic vectors and dependency trees to guide the model generate empathetic responses.
11) DCKS\citep{cai2023improving}: A model uses adaptive module for commonsense knowledge selection to ensure consistency between the model responses and the history context.
12) CTSM\citep{yufeng2024ctsm}: A model that categorizes emotions into fine-grained trait and state emotion to enhance the ability of perceiving sentiment.

\section{Empathetic Response Derived from Different Prompts}
\label{llm-prompt}

To maximize the empathetic response capabilities of the model, we developed several iterations of prompts. In practice, given the same contextual information, we assessing the impact of various prompt based on GPT-4-0613. The detailed comparative results is shown in Table~\ref{tab:llm_eval_prompt}. 

\section{Details of Human Evaluation}
\label{sec:human eva}

We apply human evaluation to assess the Coherence(\textbf{Coh.}), Empathy(\textbf{Emp.}), Informativeness(\textbf{Inf.}) and Continuity(\textbf{Cont.}) of responses from empathetic models.

For each evaluation indicator, we specify a scoring dimension from 1 to 5, as follows:

Rating 1: The generated response completely fails to meet the specific requirement.

Rating 2: The response generally meets the specific requirements, but has some shortcomings.

Rating 3: The response does meet the required requirements.

Rating 4: The response meets specific requirements while also making some additional contribution or merit.

Rating 5: The response exceeds expectations and provides excellent content or solutions while meeting specific requirements.

\section{Codes of Compared Method}
\label{sec:compared method}
We use PyTorch to implement our model. The encoder and decoder are from base version of BART as same as \cite{cai2023improving}. The CoNECT data are derived from LLaMA2-13b-Chat model on NVIDIA-A100 GPU. Besides, we train the empathetic response model Lamb using Adam optimizer with initial learning rate 0.00005 in 5 epochs and the batch size is set to 16.
All empathetic response experiments is conducted on NVIDIA-4090 GPU and Intel(R) Xeon(R) Gold 6338 CPU @ 2.00GHz. 
Qwen1.5-MoE-A2.7B-Chat, LLaMA2-13B-Instruct, LLaMA3-8B-Instruct and Mixtral-8x7B-Chat are constructed based on vLLM-v0.3.2 and sampling temperature is 0.8, top-p is 0.95.
We use the official DCKS code and reproduced it in our experimental environment. See the Appendix~\ref{sec:compared method} for detailed code links.

The existing offical codes that are involved in this work include: 
DCKS\footnote{\url{https://github.com/Hanscal/DCKS}}, 
CAB\footnote{\url{https://github.com/geri-emp/CAB}}, 
KEMP\footnote{\url{https://github.com/qtli/KEMP}}, 
CASE\footnote{\url{https://github.com/jfzhouyoo/case}}, 
SEEK\footnote{\url{https://github.com/wlr737/EMNLP2022-SEEK}}, 
CEM\footnote{\url{https://github.com/Sahandfer/CEM}}.
The evaluation metrics from Natural Language Toolkit\cite{bird2009analyzing}, same as \citet{cai2023improving}.


\begin{table*}[htbp]\small
  \centering  
  \caption{An example of EMPATHETICDIALOG and the responses generated from Lamb and other SOTA models. \textcolor{blue}{Blue text} and \textcolor{orange}{orange text} denotes the rational and emotional expression separately.}
  \resizebox{\linewidth}{!}{
    \begin{tabular}{p{4.415em}|p{30.665em}}
    \toprule
    \multirow{3}[2]{*}{\textbf{Context}} & \textbf{Speaker:} \textcolor{blue}{One of the times i remember feeling the most blissed out in life was right after the birth of my first child.} \\
    \multicolumn{1}{l|}{} & \textbf{Listener:} \textcolor{blue}{That is a very blessed day. It is something you will never forget.} \\
    \multicolumn{1}{l|}{} & \textbf{Speaker:} \textcolor{orange}{Obviously there are the demands of a new child -- but that feeling of finally meeting someone you waited so long for, and the love surrounding the whole situation. truly something to remember.} \\
    \midrule
    \textbf{Emotion} & Grateful \\
    \textbf{CoNECT} & Based on the content of the dialogue, it is clear that the \textbf{speaker} is expressing feelings of gratitude and contentment. The use of phrases such as "\textcolor{blue}{most blissed out in life}" and "\textcolor{blue}{something you will never forget}" suggest a strong \textbf{positive emotion}. Additionally, the speaker mentions the birth of their \textbf{first child}, which is a significant life event that can evoke powerful emotions. The sentiment label "grateful" accurately captures the speaker's emotional state. Overall, the causal link between the dialogue content and the sentiment label is strong and straightforward. \\
    \midrule
    \textbf{KEMP}  & I am sure you will do great !   \\
    \textbf{SEEK}  & That is a great accomplishment ! \\
    \textbf{DCKS}  & That is great to hear. i am happy for you. \\
    \midrule
    \textbf{LLaMA2-13B-Chat}  & Oh my gosh, I can totally relate! Having my first child was one of the most surreal and magical experiences of my life. The pure joy and love I felt in that moment was indescribable. It's like nothing else mattered except for that little bundle of happiness in my arms. \\
    \midrule
    \textbf{Lamb}  & Yes, it is a great feeling to finally \textcolor{purple}{meet someone} you \textcolor{purple}{waited so long} for. \\
    \midrule
    \textbf{Gold}  & Yes, I could not agree more. it is remarkable how your feeling suddenly change. \\
    \bottomrule
    \end{tabular}%
    }
  \label{tab:main_case}%
\end{table*}%

\begin{table*}[htbp]
  \centering
  \caption{Empathetic response case-2. }
  \resizebox{\linewidth}{!}{
    \begin{tabular}{p{4.415em}|p{30.665em}}
    \toprule
    \multicolumn{1}{l|}{\multirow{5}[2]{*}{\textbf{Context}}} & \multicolumn{1}{l}{\textbf{Speaker:} Hi, I feel so lonely sometimes because all my friends live in} \\
    \multicolumn{1}{l|}{} & \multicolumn{1}{l}{         a different country} \\
    \multicolumn{1}{l|}{} & \multicolumn{1}{l}{\textbf{Listener:} Oh, I am sure you are lonely. Maybe you can join some kind of} \\
    \multicolumn{1}{l|}{} & \multicolumn{1}{l}{          club that lets you meet new friends?} \\
    \multicolumn{1}{l|}{} & \textbf{Speaker:} I was thinking about it! I wanted to join a group for local moms \\
    \midrule
    \textbf{Emotion} & Lonely \\
    \textbf{CoNECT} & Based on the dialogue content, the speaker is expressing a feeling of loneliness due to their friends living in a different country. The sentiment label is questioning, as the speaker is seeking suggestions for meeting new friends in their local area. The causal link between the dialogue content and the sentiment label is that the speaker's feeling of loneliness is driving their desire to seek out new social connections. \\
    
    \midrule
    \textbf{KEMP} & I am sorry to hear that . I hope you can find a new friends !   \\
    \textbf{SEEK} & I am sure they are going to be a good friend .  \\
    \textbf{DCKS} & I am sure you will have a great time. \\
    \midrule
    \textbf{LLaMA2-13B-Chat} & Oh, that sounds like a great idea! There are probably a lot of other moms in the same situation as you who would love to meet new friends too. Have you tried reaching out to any local mom groups or clubs in your area? They might have meetups or events where you can connect with other moms and make some new friends. \\
    \midrule
    \textbf{Lamb} & That is a good idea. Maybe you could see if you could join some sort of club? \\
    \midrule
    \textbf{Gold} & That is a good idea ! This way you can also meet friends for yourself , but also maybe meet new friend 's for your children to hang out with while you do with their moms ! \\
    \bottomrule
    \end{tabular}%
    }
  \label{tab:appendix_case1}%
\end{table*}%

\begin{table*}[htbp]
  \centering
  \caption{Empathetic response case-3.}
  \resizebox{\linewidth}{!}{
    \begin{tabular}{p{4.415em}|p{30.665em}}
    \toprule
    \textbf{Context} & \textbf{Speaker:} I am so sad because I have read an article about a newborn girl who died because her parents did not believe in medication and doctors \\
    \midrule
    \textbf{Emotion} & Sad \\
    \midrule
    \textbf{KEMP}  & I am so sorry to hear that . I hope you can find a way to get a new one ? \\
    \textbf{SEEK}  & Oh wow , that is a lot of people .  \\
    \textbf{DCKS}  & I am sorry to hear that. \\
    \midrule
    \textbf{LLaMA2-13B-Chat}  & Oh no, that's heartbreaking! I can't imagine how devastating it must be for the parents to lose their child like that. It's understandable that you're feeling sad after reading that. Can I give you a hug? \\
    \midrule
    \textbf{Lamb}  & \multicolumn{1}{l}{That is so sad. I am sorry you had to go through that.} \\
    \midrule
    \textbf{Gold}  & Ugh , those articles always get me too ... : ( what was wrong with her ? \\
    \bottomrule
    \end{tabular}%
    }
  \label{tab:appendix_case2}%
\end{table*}%

\begin{table*}[htbp]
  \centering
  \caption{The effect of different versions of prompt on the empathetic response of LLM generation. \textcolor{red}{Red text} represent keywords of prompt.}
  \resizebox{\linewidth}{!}{
    \begin{tabular}{p{21.165em}p{24.085em}}
    \toprule
    \multicolumn{2}{c}{\textbf{Context}} \\
    \midrule
    \multicolumn{2}{p{45.25em}}{Speaker: i cheated on my girlfriend recently . i feel horrible about myself right now .\newline{}Listener:   } \\
    \midrule
    \multicolumn{1}{c}{\textbf{Prompt}} & \multicolumn{1}{c}{\textbf{Response}} \\
    \midrule
    You are an expert listener with empathy. Please give an appropriate response based on the provided conversation history.\textbackslash{}n\textbackslash{}nspeaker: i cheated on my girlfriend recently . i feel horrible about myself right now .\textbackslash{}nlistener:" & It's important to take responsibility for your actions and acknowledge the pain and hurt that you've caused. Consider apologizing to your girlfriend and taking steps to make amends, such as being transparent about your actions, being honest about what led to the infidelity, and making a commitment to be faithful in the future. \\
    \midrule
    \multicolumn{1}{c}{\textbf{Prompt}} & \multicolumn{1}{c}{\textbf{Response}} \\
    \midrule
    You are an expert listener with empathy. Please give an appropriate response based on the provided conversation history. \textcolor{red}{Remember, the length of the response should match the scenario of daily conversation.}\textbackslash{}n\textbackslash{}nspeaker: i cheated on my girlfriend recently . i feel horrible about myself right now .\textbackslash{}nlistener: & It's also crucial to reflect on why you cheated and address any underlying issues that may have contributed to your behavior. This could involve seeking therapy or counseling to work through any emotional or psychological issues, as well as communicating with your partner about what you need to change in order to rebuild trust and move forward. \\
    \midrule
    \multicolumn{1}{c}{\textbf{Prompt}} & \multicolumn{1}{c}{\textbf{Response}} \\
    \midrule
    You are an expert listener with empathy. Please give an appropriate response based on the provided conversation history. Remember, the length of the response should match the scenario of daily conversation \textcolor{red}{that usually is composed by 2 or 3 sentences.}\textbackslash{}n\textbackslash{}nspeaker: i cheated on my girlfriend recently . i feel horrible about myself right now .\textbackslash{}nlistener: & Remember that forgiveness is a process and it may take time for your girlfriend to heal and trust you again. Be patient and supportive, and consider taking concrete steps to show her that you're committed to making things right. Ultimately, it's up to your girlfriend to decide whether she can forgive you, but by taking responsibility and working to make things right, you can at least demonstrate your sincerity and commitment to the relationship. \\
    \midrule
    \multicolumn{1}{c}{\textbf{Prompt}} & \multicolumn{1}{c}{\textbf{Response}} \\
    \midrule
    You are a professional psychologist. Please give a reply from the perspective of a friend based on the patient's situation. \textcolor{red}{Be careful to avoid a large number of professional vocabulary in your reply, and the expression should be natural.}\textbackslash{}nThe following is a set of historical conversations between you and your best friend. Please respond according to the context and pay attention to fully consider the emotional information in the historical conversations when replying.\textbackslash{}n\textbackslash{}nspeaker: i cheated on my girlfriend recently . i feel horrible about myself right now .\textbackslash{}nlistener: & Hey buddy, it sounds like you're going through a tough time. Nothing I say can change what happened, but I'm here for you. You did something you regret, it's plain to see that you're feeling guilty, which shows you've realised your mistake. It's important now to think about what you want to do moving forward and how you can avoid repeating this in future. Consider coming clean with your girlfriend if it feels right. Sometimes, owning up to our mistakes and facing the consequences can actually help us grow as individuals. But remember, whatever you decide, I'm here for you, man. \\
    \midrule
    \multicolumn{1}{c}{\textbf{Prompt}} & \multicolumn{1}{c}{\textbf{Response}} \\
    \midrule
    You are a professional psychologist. Please give a reply from the perspective of a friend based on the patient's situation. Be careful to avoid a large number of professional vocabulary in your reply, and the expression should be natural.\textbackslash{}nThe following is a set of historical conversations between you and your best friend. Please respond according to the context and pay attention to fully consider the emotional information in the historical conversations. Remember, response should not exceed 30 words.\textbackslash{}n\textbackslash{}nSpeaker: i cheated on my girlfriend recently . i feel horrible about myself right now .\textbackslash{}nListener:  & I'm really sorry to hear that you're going through this. It's important to confront your feelings and be honest with her about it. \\
    \bottomrule
    \end{tabular}%
    }
  \label{tab:llm_eval_prompt}%
\end{table*}%

\end{document}